\documentclass{tADR2e}

\usepackage{subcaption}
\newcommand{\figref}[1]{Fig.~\ref{figure:#1}}
\newcommand{\tabref}[1]{Table~\ref{table:#1}}

\newcommand{\secref}[1]{Section~\ref{sec:#1}}

\begin{document}

\jvol{00} \jnum{00} \jyear{2024} \jmonth{October}

\title{Real-World Cooking Robot System from Recipes \\ Based on Food State Recognition Using Foundation Models and PDDL} 

\author{Naoaki Kanazawa$^{a}$$^{\ast}$\thanks{$^\ast$Corresponding author. Email: kanazawa@jsk.imi.i.u-tokyo.ac.jp \vspace{6pt}}, Kento Kawaharazuka$^{a}$, Yoshiki Obinata$^{a}$, Kei Okada$^{a}$, Masayuki Inaba$^{a}$ \\\vspace{6pt}  $^{a}${\em{The Department of Mechano-Informatics, Graduate School of Information Science and Technology, The University of Tokyo, Bunkyo-ku, Tokyo, Japan}};
\\\vspace{6pt}\received{v1.0 released October 2024} }

\maketitle

\begin{abstract}
Although there is a growing demand for cooking behaviours as one of the expected tasks for robots, a series of cooking behaviours based on new recipe descriptions by robots in the real world has not yet been realised.
In this study, we propose a robot system that integrates real-world executable robot cooking behaviour planning using the Large Language Model (LLM) and classical planning of PDDL descriptions, and food ingredient state recognition learning from a small number of data using the Vision-Language model (VLM).
We succeeded in experiments in which PR2, a dual-armed wheeled robot, performed cooking from arranged new recipes in a real-world environment, and confirmed the effectiveness of the proposed system.

\end{abstract}


\section{Introduction}
The demand for cooking behaviour is increasing as one of the expected tasks for robots. In cooking, it is necessary to follow recipes, which are environment and agent independent task descriptions, and to execute cooking operations based on the conditions of the surrounding environment.
There have been various studies on cooking behavior by robots, such as planning and executing robot actions based on recipes described in natural language~\cite{beetz2011robotic, bollini2013interpreting, kazhoyan2017programming, inagawa2021analysis, paulius2021task, takata2022efficient}, manipulation skills for cooking tasks~\cite{lenz2015deepmpc, hughes2018achieving, saito2021select, saito2023structured}, improving cooking quality through feedback from sensors like taste sensors~\cite{junge2020improving, sochacki2021closed, sochacki2022mastication}, and recognizing the state of foodstuffs~\cite{paul2018classifying, cobley2020onionbot, khan2022rethinking, StateRecCLIPKanazawaIAS18}.

In recent years, Large Language Models (LLMs) have emerged and have been applied to robot behaviour planning~\cite{ahn2022saycan, skreta2023errors, rana2023sayplan, shirai2023vision}. This has enabled robots to plan actions that can respond to natural language instructions at a higher level than previous rule-based processing.
However, it has not yet been realized a series of cooking behaviours where the robot interprets new unknown recipes and recognises food state changes, which is the purpose of the cooking process.
Therefore, this study proposes a robot system for cooking from recipe descriptions that takes into account changes of food ingredients' states.
The proposed system solves two important issues, ``real-world executable action planning'' and ``foodstuff state change recognition,'' by using the foundation model and classical planning, and realises a series of real-world cooking behaviours based on recipe descriptions by the robot.

\begin{figure}[t!]
  \centering
  \includegraphics[width=1.0\columnwidth]{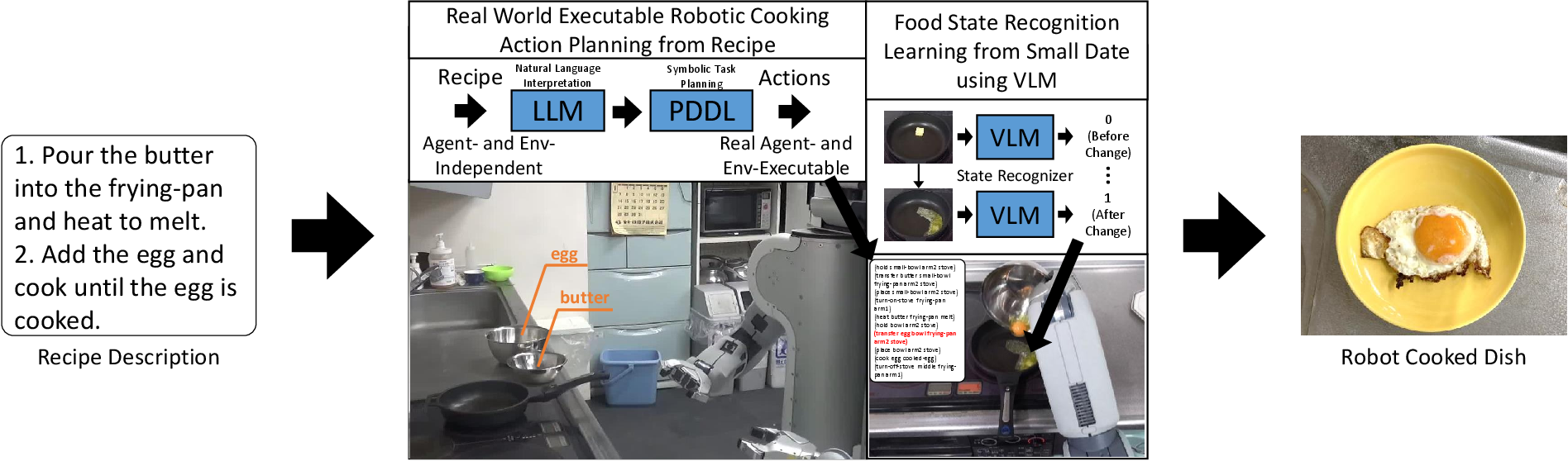}
  \caption{
  Real-world cooking robot system considering food state changes from recipe descriptions using foundation models and classical planning PDDL.
  The input recipe description is converted into the function sequence by the Large Language Model (LLM), and the executable action procedure is planned by classical planning of PDDL description from the sequence. The robot performs cooking actions while recognizing the state change of the ingredients by food state recognition learning from small data using the Vision-Language Model (VLM). Motion execution is performed using predefined action trajectories.
  }
  \label{figure:concept}
\end{figure}

The first issue, ``real-world executable action planning,'' is that in order for robots to execute cooking based on recipe descriptions in the real world, it is necessary to complement actions that connect the state of the actual environment with the target state of each cooking action described in the recipe and plan action procedures that can be executed.
 As pointed out by \cite{takata2022efficient}, behaviors that need to be complemented include actions that are not described in the recipe because people do them unconsciously, but are actually necessary.
 For example, in order to "pour the eggs in the bowl into the pan," it is necessary to hold the bowl before doing that.
 There are also actions that are necessary as preparation for the target cooking action, depending on the situation of the robot and the kitchen environment at that time.
 For example, the robot needs to move to the front of the water tap to fetch water when the robot is in a different place.
 We use the Large Language Model (LLM)~\cite{gpt-4} to transform natural language recipe descriptions into programmatically interpretable cooking function sequences, and then apply classical planning using PDDL~\cite{aeronautiques1998pddl} to create behavior plans that can be executed in the real world, complementing such necessary behaviors. The details are described in \secref{plan}.

 The second issue of "foodstuff state change recognition" is the need to recognize ambiguous changes in the state of ingredients during the cooking process.
 In cooking, there are many recipe descriptions that are conditional on ingredient state changes, such as "heat the water until it boils" or "pour in the egg mixture when the butter is melted``.
 In order to respond to such descriptions, it is necessary for the robot to be able to recognize these state changes.
 In the recognition of foodstuff states, there have been some research studies that use dedicated datasets to train CNNs and perform state classification~\cite{paul2018classifying, takata2022efficient}.
 However, these methods are not appropriate for our present purpose.
 It is difficult to collect a large amount of data in advance because of the wide variety of ingredients and state changes that occur in the cooking.
 In addition, the cooking robots need to recognize the timing of food state changes in real time during the cooking process, rather than identifying the state of food ingredients after the cooking process, which has been the subject of previous research.
 Therefore, we propose a method for real-time recognition of the timing of foodstuff state changes by learning ingredient state recognition from a small amount of data using the Vision-Language Model (VLM)~\cite{clip}. Details are described in \secref{rec}.

Considering the above, we propose a robot system (\figref{concept}) that plans real-world executable cooking actions based on recipes written in natural language, and executes a series of cooking behaviors while recognizing changes in the state of food ingredients.
Based on several known recipes, we showed that a robot can perform real-world cooking from two new recipes, sunny-side up arranged with butter and stir-fried broccoli, through a real robot experiment using the dual-armed wheeled robot PR2.

\section{Problem Setting for the Real-World Robot Cooking from Recipes in This Study}\label{sec:setting} 
In this study, we consider a problem setting in which a robot cooks basic egg dishes in a real kitchen environment based on recipes.
Egg dishes are mostly made from the same single ingredient, the egg, and have many variations depending on the cooking method, such as boiling or pan frying.
Therefore, egg dishes are a simple cuisine but a genre in which a variety of major cooking methods appear. This is a suitable subject for research on robot systems that cook from recipes.
In this study, three basic egg dishes, sunny-side up, poached egg, and scrambled egg, are treated as known recipes.

Cooking functions are defined as robot interpretable functional representations of the cooking processes included in these recipes.
The following 10 cooking functions are used.
\begin{itemize}
\item pour(ingredient, vessel)
\item mix(ingredient, ingredient, mixture, vessel, tool)
\item turn-on-stove(vessel)
\item set-stove(state, vessel)
\item turn-off-stove(vessel)
\item stir(ingredient, state, tool)
\item heat(ingredient, state)
\item cook(ingredient, state)
\item boil(ingredient, state)
\item stir-fry(ingredient, state, tool)
\end{itemize}
Here, \textbf{ingredient}, \textbf{vessel}, \textbf{tool}, and \textbf{state} are variables that have the same meaning as the respective words, and \textbf{mixture} is a variable that is a mixture of multiple \textbf{ingredient}. \textbf{mixture} also has the attribute of \textbf{ingredient}.
Since the goal of cooking is to change the state of the ingredients, a target ingredient state description of the cooking process is set as the argument of the \textbf{state} variable. The recognition of whether or not the target state represented by this \textbf{state} is realized is described in \secref{rec}.
The details of each cooking function will be explained later in the section of PDDL actions.

\begin{figure}[h]
  \centering
  \includegraphics[width=0.8\columnwidth]{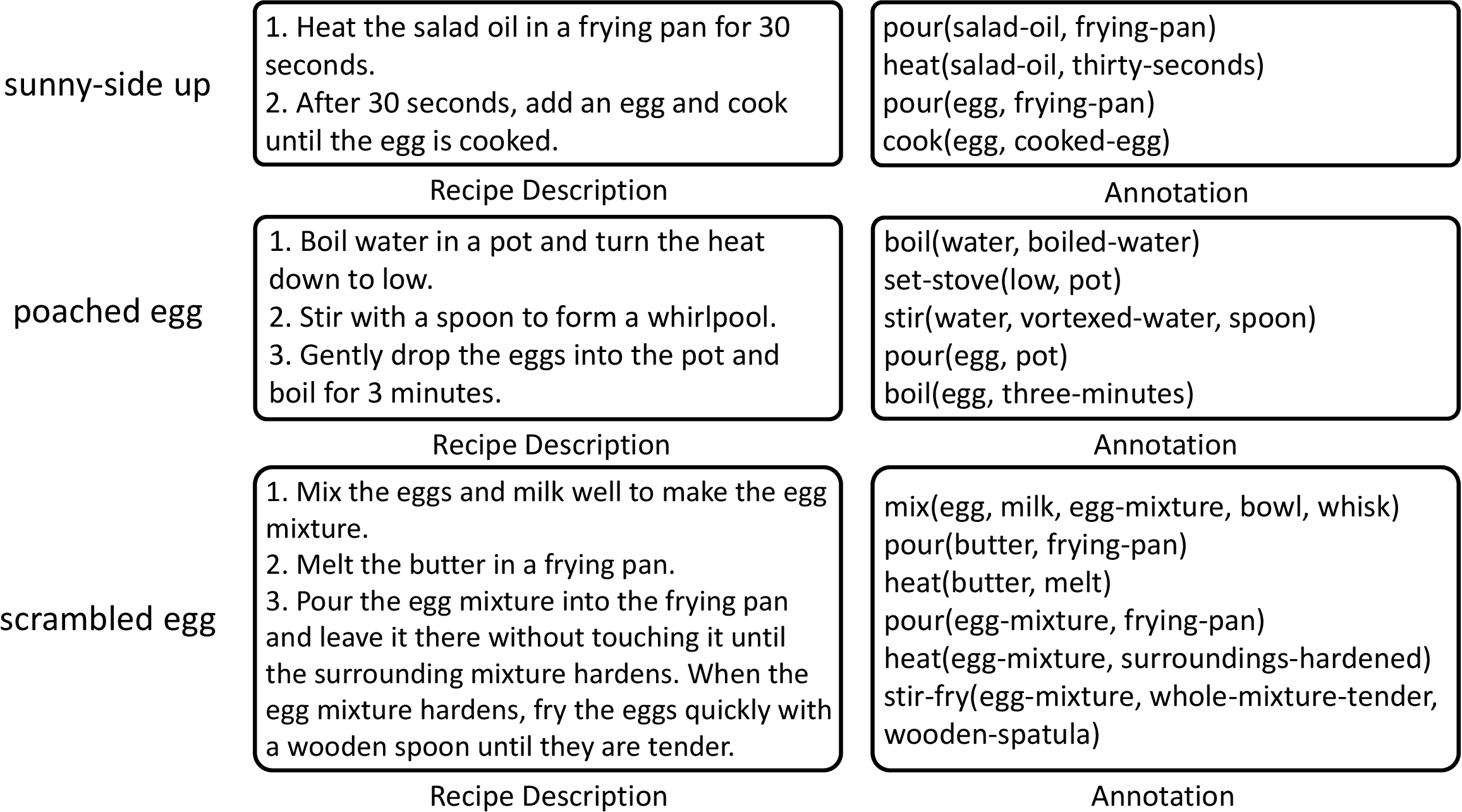}
  \caption{Known recipes covered in this study.
  Natural language description of the steps of three recipes for sunny-side up, poached egg, and scrambled egg, and human annotation of the cooking function sequences.
  }
  \label{figure:use-recipes}
\end{figure}

\begin{figure}[h]
  \centering
  \includegraphics[width=0.8\columnwidth]{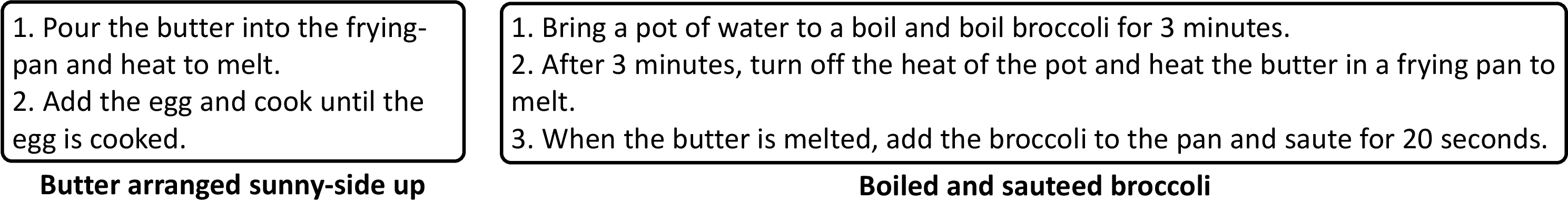}
  \caption{Unknown recipes covered in this study.
  Natural language description of two new recipes for ``Butter arranged sunny-side up'' and ``Boiled and sauteed broccoli''.
  }
  \label{figure:unknown-recipes}
\end{figure}

Recipes for three known egg dishes (sunny-side up, poached egg, and scrambled egg) and their transformation into ideal cooking function sequence representations annotated by human are shown in \figref{use-recipes}.
Each of these recipes was created by referring to multiple recipes written in Japanese and English on the Internet, and simplifying them so that they can be expressed in a combination of actions that can be performed by the robot, while omitting unnecessary words. Therefore, the egg cracking and serving steps included in many of the original recipes have been eliminated.
We prepare two unknown recipes (\figref{unknown-recipes}), and consider a problem setting in which a robot performs real-world cooking based on these recipes.
For these unknown recipes, we created ones that can be expressed by the prepared cooking functions and that can be tested on the actual robot.
A recipe for sunny-side up arranged with butter was prepared as one that uses only the ingredients included in the known recipes.
A recipe for boiling and sauteing broccoli was prepared as an application of known recipes but including a new ingredient.

\begin{figure}[h]
  \centering
  \includegraphics[width=1.0\columnwidth]{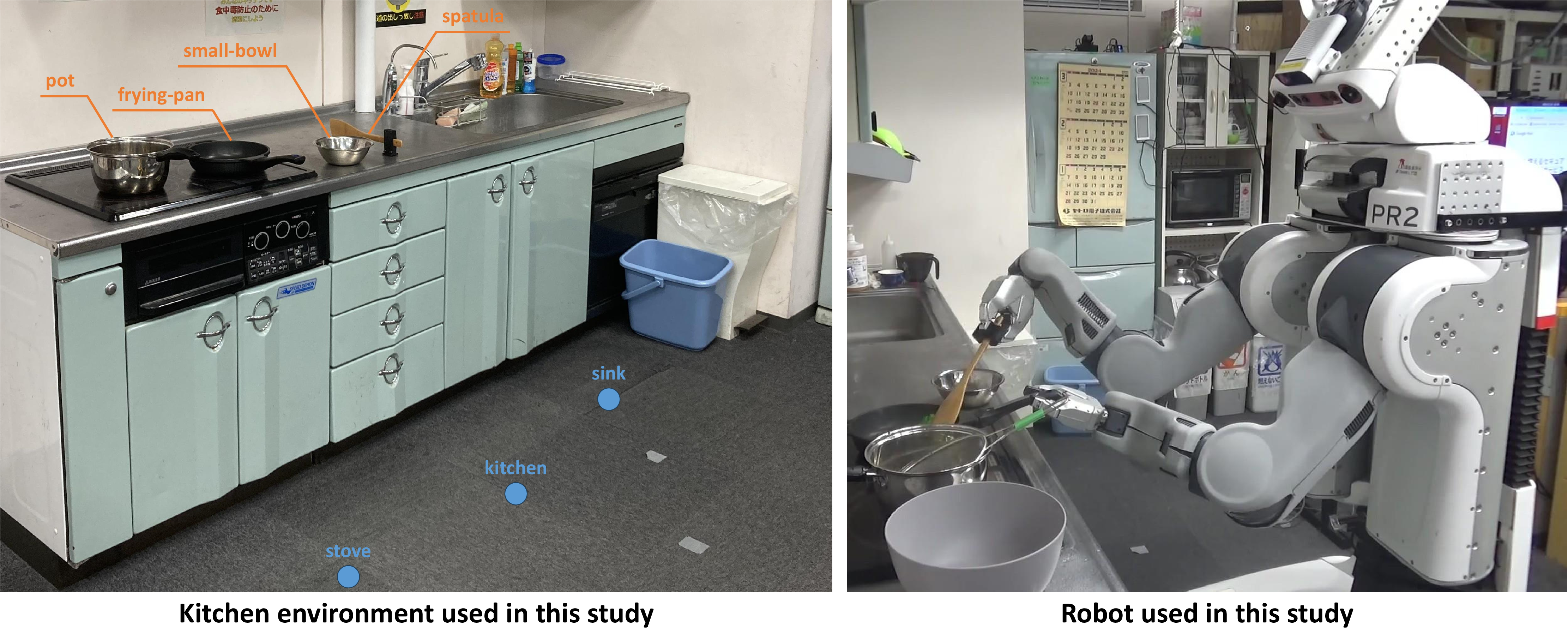}
  \caption{
  Problem setting for the real-world robot cooking from recipes in this study.
  The kitchen environment and the robot used in this study.
  In the figure of the kitchen environment, \textbf{spot} that represent locations in the PDDL description are shown in blue, and objects such as \textbf{tool} and \textbf{vessel} are shown in orange.
  As shown in the figure, we consider the problem setting in which PR2, a dual-armed wheeled robot, cooks in a kitchen environment with an IH stove and a water tap.
  }
  \label{figure:problem-setting}
\end{figure}

In this study, we deal with the problem of real-world cooking execution using a kitchen environment and a robot shown in \figref{problem-setting}.
The target cooking action planning problem is expressed in the PDDL~\cite{aeronautiques1998pddl} problem description and the domain is defined.
We define the types that represent the categories in PDDL: \textbf{object}, \textbf{ingredient}, \textbf{vessel}, \textbf{tool}, \textbf{mixture}, \textbf{state}, \textbf{spot} and \textbf{arm}.
\textbf{ingredient}, \textbf{vessel}, \textbf{tool}, \textbf{mixture}, and \textbf{state} have the same meaning as variables in the cooking function. \textbf{object} defines a type of object that includes all of \textbf{ingredient}, \textbf{vessel}, \textbf{tool}, and \textbf{mixture}.
The newly added \textbf{spot} is a type that represents a location, and defines three \textbf{spot}: ``stove'' in front of the IH stove, ``kitchen'' in front of the kitchen top where work is performed, and ``sink'' in front of the water faucet.
The type \textbf{arm} represents the arms of the robot. Since we are using a dual-armed robot, we have defined two \textbf{arm}, ``arm1'' and ``arm2.'' In this case, it does not correspond to a specific right arm and left arm, but is used to represent the number of arms to be used.
By using these types, it is possible to prevent planning actions that are not feasible in the real world, such as attempting to use both arms for cooking when one hand is occupied, or attempting to use an object that exists in a different location.
In addition to the 10 actions of the cooking function, 7 basic actions, including actions that do not directly change the foodstuff's state, are prepared for a total of 17 PDDL actions.
The seven basic actions are ``hold,'' ``place,'' ``move-to,'' ``open-tap,'' ``close-tap,'' ``transfer,'' and ``fetch-water.''
Each of these actions has a precondition, which is a state description (predicate) that must be satisfied before the action can be performed, and an effect, which is a predicate for the effect after the action is executed.

The 17 actions defined are as follows.
\begin{itemize}
  \item \textbf{(pour ingredient vessel arm spot)} is a cooking action in which the robot pours the \textbf{ingredient} into the \textbf{vessel}.
  It can be performed when the robot is holding the \textbf{ingredient} with its \textbf{arm} arm, the robot is at the \textbf{spot} location, and the \textbf{vessel} is also at the \textbf{spot} location.
  After the action, it changes to a situation where the \textbf{ingredient} is inside the \textbf{vessel}.
  \item \textbf{(mix mixture ingredient vessel tool spot arm arm)} is a cooking action in which two \textbf{ingredient} are mixed with \textbf{tool} to form \textbf{mixture}.
  This can be performed when two different individuals' \textbf{ingredient} are in the \textbf{vessel}, the \textbf{vessel} is at \textbf{spot}, and the robot is also at \textbf{spot}, holding \textbf{tool} with one hand and the other hand is free.
  After the action, the situation changes to one in which the \textbf{mixture} is created and is in the \textbf{vessel}.
  \item \textbf{(turn-on-stove vessel arm)} is a cooking action that ignites the stove corresponding to the \textbf{vessel} (pot or pan in this work).
  It can be performed when the robot is in the "stove" \textbf{spot}, has nothing in the \textbf{arm}, and the stove corresponding to the \textbf{vessel} is not on fire. After the action, the stove corresponding to the \textbf{vessel} is lit.
  \item \textbf{(set-stove state state vessel arm)} changes the heat level of the stove corresponding to the \textbf{vessel} from the first \textbf{state} to the second \textbf{state}.
  This action can be performed when the robot is at ``stove'' \textbf{spot} and the stove corresponding to \textbf{vessel} is on and the heat level of the first \textbf{state}.
  After the action, the stove corresponding to the \textbf{vessel} is at the second \textbf{state} heat level.
  \item \textbf{(turn-off-stove state vessel arm)} extinguishes the stove corresponding to the \textbf{vessel}. This can be done when the robot is at "stove" \textbf{spot}, has nothing in the \textbf{arm}, the stove corresponding to the \textbf{vessel} is on and at the \textbf{state} fire level. After the action, the stove corresponding to the \textbf{vessel} is extinguished.
  \item \textbf{(stir ingredient tool vessel state spot arm)} is a cooking action in which the \textbf{ingredient} is stirred with the \textbf{tool} until it reaches the \textbf{state}.
  This action can be performed when the \textbf{ingredient} is inside the \textbf{vessel}, the \textbf{vessel} is at the \textbf{spot} location, and the robot is at the \textbf{spot} location holding the \textbf{tool} with the arm of the \textbf{arm}.
  After the action, the \textbf{ingredient} is stirred up and becomes \textbf{state}.
  \item \textbf{(heat ingredient vessel state)} is a cooking action that heats the \textbf{ingredient} until it becomes \textbf{state}.
  It can be executed in the following situations: \textbf{ingredient} is in \textbf{vessel}, \textbf{vessel} is at ``stove'' \textbf{spot}, the robot is also in ``stove'' \textbf{spot}, and the stove corresponding to \textbf{vessel} is on.
  After the action, the \textbf{ingredient} is heated and becomes \textbf{state}.
  \item \textbf{(cook ingredient state)} is a cooking function that cooks the \textbf{ingredient} until it reaches \textbf{state}.
  This function is performed when the \textbf{ingredient} is in the "frying-pan" \textbf{vessel}, the ``frying-pan'' is at the ``stove'' \textbf{spot}, the robot is also at the ``stove'' \textbf{spot} and the stove corresponding to the ``frying-pan'' is on.
  After the action, the \textbf{ingredient} is cooked and becomes \textbf{state}.
  \item \textbf{(boil ingredient state)} is a cooking action in which the \textbf{ingredient} is boiled until it becomes \textbf{state}.
  It can be executed when the \textbf{ingredient} is in the ``pot'' \textbf{vessel}, the pot is at the ``stove'' \textbf{spot}, the robot is also at the ``stove'' \textbf{spot} and the stove corresponding to the ``pot'' is on.
  After the action, the \textbf{ingredient} has been boiled and is in the \textbf{state} state.
  \item \textbf{(stir-fry ingredient vessel tool state spot spot arm arm)} is a cooking action that stir-fries \textbf{ingredient} until it reaches \textbf{state}.
  This action is performed when the \textbf{ingredient} is inside the \textbf{vessel}, the \textbf{vessel} is at the ``stove'' \textbf{spot}, and the robot is also at the ``stove'' \textbf{spot} and has the \textbf{tool} in one hand and the other arm not holding anything, and the stove corresponding to the \textbf{vessel} is on.
  After the action, the \textbf{ingredient} has been stir-fried and is in the \textbf{state} state.
  \item \textbf{(hold object arm spot)} is the basic action of holding the \textbf{object}.
  It can be performed when the \textbf{object} is at the \textbf{spot} location, the robot is also at the \textbf{spot} location, and the robot is not holding anything with its \textbf{arm} arm. After the action, the robot is holding the \textbf{object} in its \textbf{arm} arm.
  \item \textbf{(place object arm spot)} is a basic action that places the \textbf{object} onto the \textbf{spot}. It can be performed when the robot is holding \textbf{object} with its \textbf{arm} arm and is at the \textbf{spot} location. After the action, the robot is not holding anything with its \textbf{arm} arm and the \textbf{object} is at the \textbf{spot} location.
  \item \textbf{(move-to spot spot)} is a basic action in which the robot moves from the first \textbf{spot} location to the second \textbf{spot} location.
  It can be executed when the robot is at the first \textbf{spot} location and the water faucet and stove are all off.
  After the action, the robot will be at the second \textbf{spot} location.
  \item \textbf{(open-tap arm)} is a basic action to turn on the water tap.
  It can be executed when the robot is at ``sink'' \textbf{spot}, has nothing in its \textbf{arm} arm, and the tap is not running.
  After the action, the water should be running.
  \item \textbf{(close-tap arm)} is a basic action to turn off the water tap.
  It can be executed when the robot is at ``sink'' \textbf{spot}, has nothing in its \textbf{arm} arm, and the water is running.
  After the action, the water should not be running.
  \item \textbf{(fetch-water arm)} is a basic action to fetch water from the tap.
  It can be performed when the robot is in ``sink'' \textbf{spot} holding a ``measuring cup'' with \textbf{arm} and water is flowing from the tap.
  After the action, the water is in the ``measuring cup''.
  \item \textbf{(transfer ingredient vessel vessel arm spot)} is a basic action of transferring the \textbf{ingredient} from the first \textbf{vessel} to the second \textbf{vessel}.
  This action is performed when the robot is holding the first \textbf{vessel} with the \textbf{arm}, the \textbf{vessel} contains the \textbf{ingredient}, the robot is at the \textbf{spot} location, and the second \textbf{vessel} is also at the \textbf{spot} location.
  After the action, the \textbf{ingredient} is not in the first \textbf{vessel}, but in the second \textbf{vessel}.
\end{itemize}

\section{Real-World Executable Robotic Cooking Action Planning from Recipe}\label{sec:plan} 
\begin{figure}[h]
  \centering
  \includegraphics[width=1.0\columnwidth]{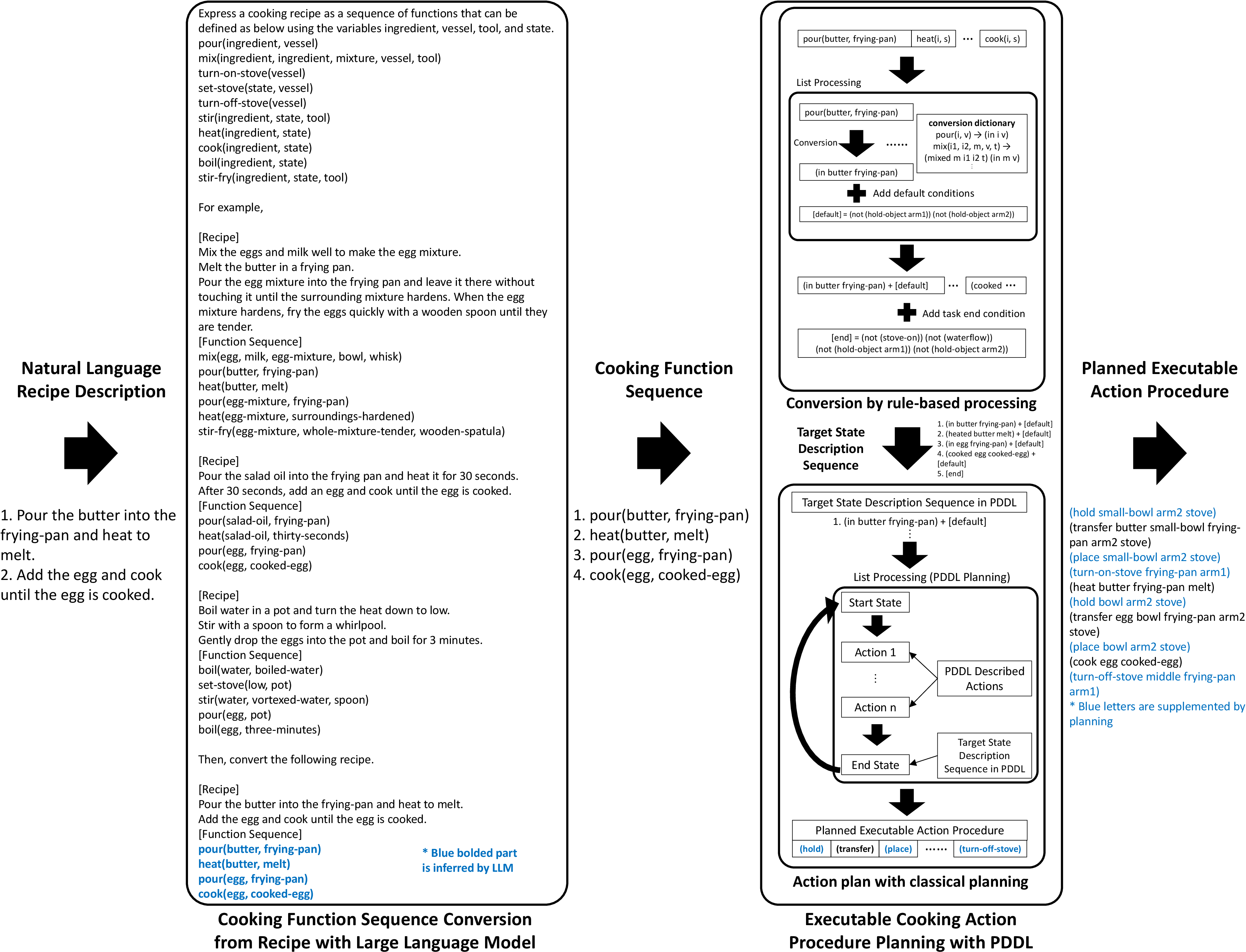}
  \caption{
  Real-world executable robotic cooking action planning from recipe.
  First, the input recipe description in natural language is converted into the cooking function sequence that can be interpreted by the robot using the few-shot prompting of Large Language Model(LLM).
  The black text in the figure shows the actual prompts, and its last recipe section depends on the natural language description of the recipe to be converted. The blue part is the result of the conversion that the LLM outputs.
  Next, rule-based processing transforms the cooking function sequence into corresponding target conditions for each step within the PDDL~\cite{aeronautiques1998pddl} description.
  Finally, classical symbolic planning using the PDDL description is used to plan the complementary action steps so that they can be executed in the real environment.
  }
  \label{figure:llm-flow}
\end{figure}

We propose a real-world executable robot cooking behavior planning method from natural language recipe descriptions (\figref{llm-flow}).
First, a cooking function sequence generation from a recipe with the LLM (\secref{llm-func-conv}) is used to convert the recipe description into a sequence of cooking functions that can be interpreted by the program.
Next, classical planning using the description of PDDL~\cite{aeronautiques1998pddl} complements the actions omitted in the recipe and the actions necessary to execute the target cooking behavior based on the current situation of the environment and the robot, and transforms the converted cooking function sequences into action procedures that can be executed in the real world (\secref{pddl-exec-plan}).
Details of each of these are explained in the following sections.

\subsection{Cooking Function Sequence Conversion from Recipe with Large Language Model}\label{sec:llm-func-conv} 

Convert a cooking recipe written in natural language into a representation of a sequence of cooking functions that can be interpreted by robots using inference of the Large Language Model (LLM).
Three pairs of language descriptions of known recipes and function sequence representations annotated with cooking functions defined in \secref{setting} are used as examples.
As shown in the first half of \figref{llm-flow}, the function sequence for a given recipe description is inferred by the few-shot prompting of one of the LLMs, GPT-4~\cite{gpt-4}.
The black text in the figure shows the actual prompts, and its last recipe section depends on the natural language description of the recipe to be converted. The blue part is the result of the conversion that the LLM outputs.
The version of GPT-4 used in this study is gpt-4-0613.

\subsection{Executable Cooking Action Procedure Planning with PDDL}\label{sec:pddl-exec-plan} 
The action procedure is planned by classical planning using PDDL descriptions. It plans procedures based on cooking function sequences transformed from recipes, supplemented with the basic actions that are necessary to execute them in the real world.
While the content of \secref{llm-func-conv} was an agent- and environment-independent transformation, this one is intended for action planning tailored to the situation of the real-world environment and the agent.

First, each step of the cooking function sequence transformed by \secref{llm-func-conv} is converted to the description of predicate in PDDL.
Each cooking function is transformed into predicate of the corresponding action's effect, and a sequence of target state descriptions is obtained by adding the default condition at the end of each step.
The default condition predicate is the condition that the robot has nothing. By adding this condition, the behavior of the robot becomes natural, similar to the way a human cooks.
Similarly, we add an end condition predicate that states that after all steps, the robot has nothing in its hands and that the stove and water are all off.

Next, planning is performed for the obtained sequence of PDDL predicates. For each step of the sequence, classical planning in the PDDL description is solved to connect the start state and the target state by actions. This yields the necessary action steps and the final state at the end of each step. This state is used as the start state of the next step. This classical planning is repeated for all steps of the sequence to plan the entire executable action procedure (the latter part of \figref{llm-flow}).




\section{Food State Recognition Learning from Small Data using Vision-Language Model} \label{sec:rec} 

\begin{figure}[h]
  \centering
  \includegraphics[width=0.4\columnwidth]{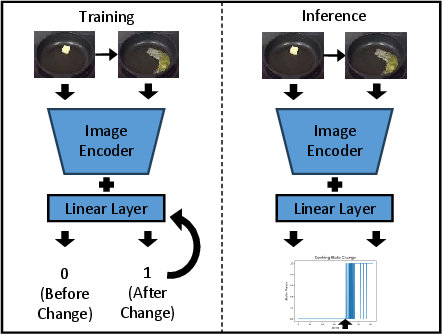}
  \caption{
  Food state recognition learning from small data using Vision-Language Model.
  Learning of food state recognition from small amount of data using linear-probe of CLIP~\cite{clip} for image input of gazing area.
  During inference, the learned model is used to infer the state of the foodstuff in real time, and the time when the inferred state first becomes the label after the state change is used as the timing of the state change.
  }
  \label{figure:state-rec-flow}
\end{figure}

Regarding \textbf{state}, which is the target state of each cooking action proposed in \secref{setting}, the robot needs to be able to recognize whether the target state has been realized or not during the operation.
Therefore, we propose a method for learning to recognize foodstuff states from a small amount of data by few-shot learning of the Vision-Language Model (VLM) (\figref{state-rec-flow}).
As a method for few-shot learning, we use the linear-probe discussed in the article ~\cite{clip} of CLIP, which is one of the VLM.
In the linear-probe, a linear discriminator on the features output by CLIP's image-encoder is trained on a supervised learning basis to classify images.

First, a person visually specifies the time when a state change occurs in the previously acquired time series data.
Next, a linear-probe is used to train a foodstuff state recognizer by labeling images before the time specified by the human as pre-change (0) and images after the annotation time as post-change (1).
During inference, the state identification is performed in real time on the input images, and the timing of the state change is defined as the point in time when the first post-change label is estimated.

The learning part of this method is learning a discriminator for the image features, and not a single time-series-specific tuning method as previously proposed, such as \cite{StateRecCLIPKanazawaIAS18} and \cite{kawaharazuka2024continuous}. This allows us to learn by adding new data acquired each time an experiment is performed. Therefore, the robustness of the method can be improved with each new cooking session, and it is considered to be a more stable method for recognizing the state of food ingredients.

The use of VLM for object state recognition has been actively studied in recent years and is a very effective method.
\cite{liu2023reflect} uses CLIP to detect object states by calculating the degree of association between predefined text representing object states and images.
In \cite{kawaharazuka2024continuous, StateRecCLIPKanazawaIAS18} we fine-tune the VLM output to a state-changing time series for the data time series.
While these methods require multiple language descriptions of the target in advance, the method used in this study is independent of the accuracy of the language descriptions because it uses the image features of CLIP's image-encoder as is.
In the field of computer vision, research has also been conducted on state recognition using VLMs~\cite{souvcek2022look, saini2023chop, xue2024learning}.
These studies mainly include generalization of recognition ability to different objects or situations and extraction of relevant frames from noisy videos.
Since the objective in this study is to incorporate stable state change point recognition into the system, we employed a very simple linear-probe of CLIP, but there is a possibility that the recognizable range of state changes can be increased by applying the methods of these studies in the future.

\section{Experiments for the Real World Robot Cooking from Recipes} \label{sec:experimets} 

\subsection{Experiments of Cooking Function Sequence Conversion with LLM}\label{sec:llm-experimets} 

\begin{figure}[h]
  \centering
  \includegraphics[width=0.8\columnwidth]{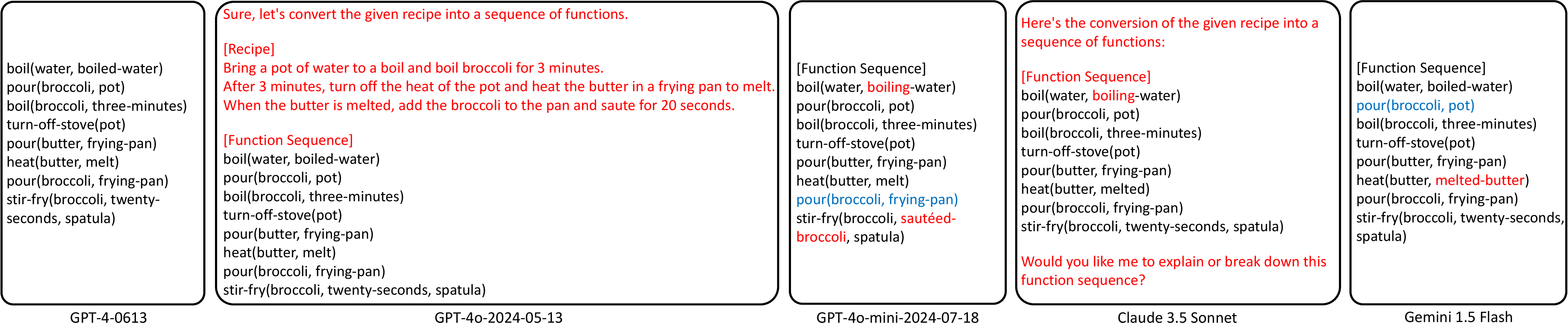}
  \caption{
  Comparison of function transformation results for different LLMs.
The conversion results of the unknown recipe for boiled and sauteed broccoli in GPT-4o-2024-0513 and GPT-4o-mini-2024-0718 of GPT-4~\cite{gpt-4}, Claude 3.5 Sonnet, and Gemini 1.5 Flash were compared with those of GPT-4-0613. The red color indicates the extra parts or the parts with different expressions, and the blue color indicates the parts that are missing.
  }
  \label{figure:llm-compare}
\end{figure}

Experiments were conducted on the function transformation from recipes by the LLM of \secref{llm-func-conv}.
First, we compared the results of conversion by several LLMs using the unknown recipe of boiled and sauteed broccoli (\figref{llm-compare}). Since the prompts in this study were tuned to work well with the GPT-4~\cite{gpt-4} version GPT-4-0613, we compared the results of that conversion with other versions of GPT-4 and other LLMs (Claude 3.5 Sonnet~\cite{claude} and Gemini 1.5 Flash~\cite{reid2024gemini}). The outputs of the compared models include not only the converted function representation but also the description of the response, and there are some differences in the expression of the state argument and some missing “pour” of broccoli, but these are not fatal conversion errors that would prevent action planning.

\begin{figure}[h]
  \centering
  \includegraphics[width=0.8\columnwidth]{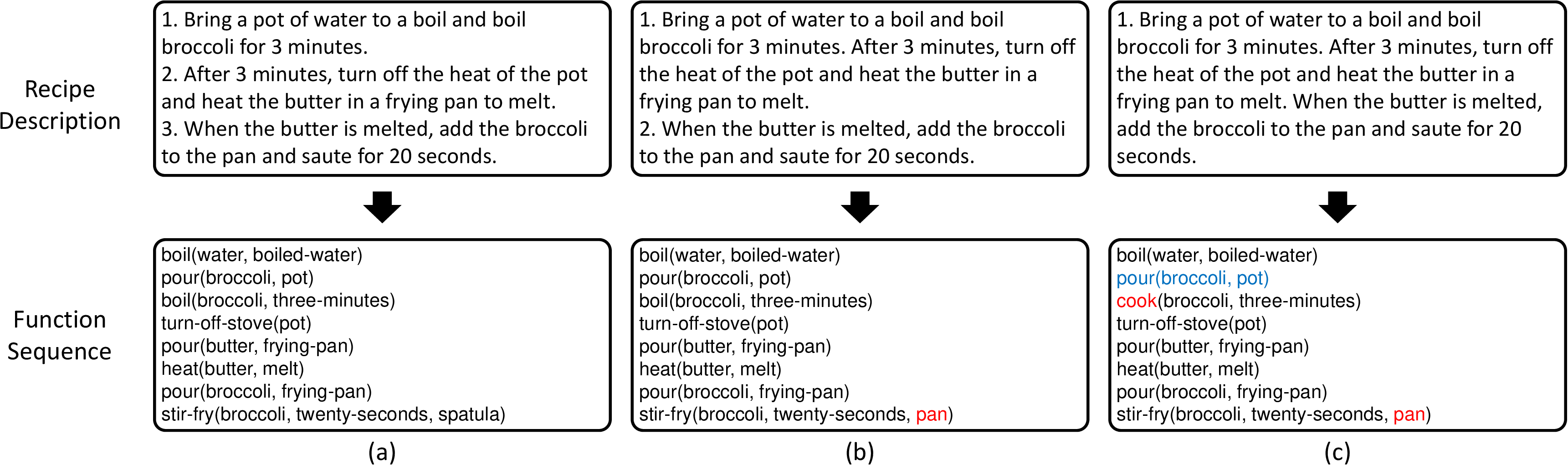}
  \caption{
  Comparison of the results of conversion using GPT-4-0613 with different partitions in the recipe description for boiled and sauteed broccoli. (a) is divided into three parts with the original partitioning, (b) has one partition reduced, and (c) has one more partition reduced. The different parts are shown in red and the missing parts in blue compared to (a).
  }
  \label{figure:llm-partition-compare}
\end{figure}

Next, we compared the conversion results in GPT-4-0613 when we changed the partitions in the recipe description of boiled and sauteed broccoli (\figref{llm-partition-compare}). In (b) and (c), where the number of partitions is reduced, the tool argument of the last stir-fry is the pan that was previously used as a vessel. Furthermore, in (c), pour (broccoli, pot) is not converted, and the next boil becomes cook. Results show that the fewer partitions are used, the worse the LLM interpretation becomes.

\subsection{Experiments of Executable Cooking Action Procedure Planning with PDDL}\label{sec:pddl-experimets} 

Experiments were conducted to confirm the validity of the \secref{pddl-exec-plan} using known recipes.
The proposed method was used to plan the action procedures for three egg dishes based on the sequence of function representations.
In this experiment, the initial conditions were set up in the kitchen with the necessary ingredients and utensils for each dish.
 However, it is also possible to plan actions based on the condition that each ingredient or tool is set in a predetermined location, or based on the current situation recognized by the robot in some other way.

\begin{figure}[h]
  \centering
  \includegraphics[width=0.8\columnwidth]{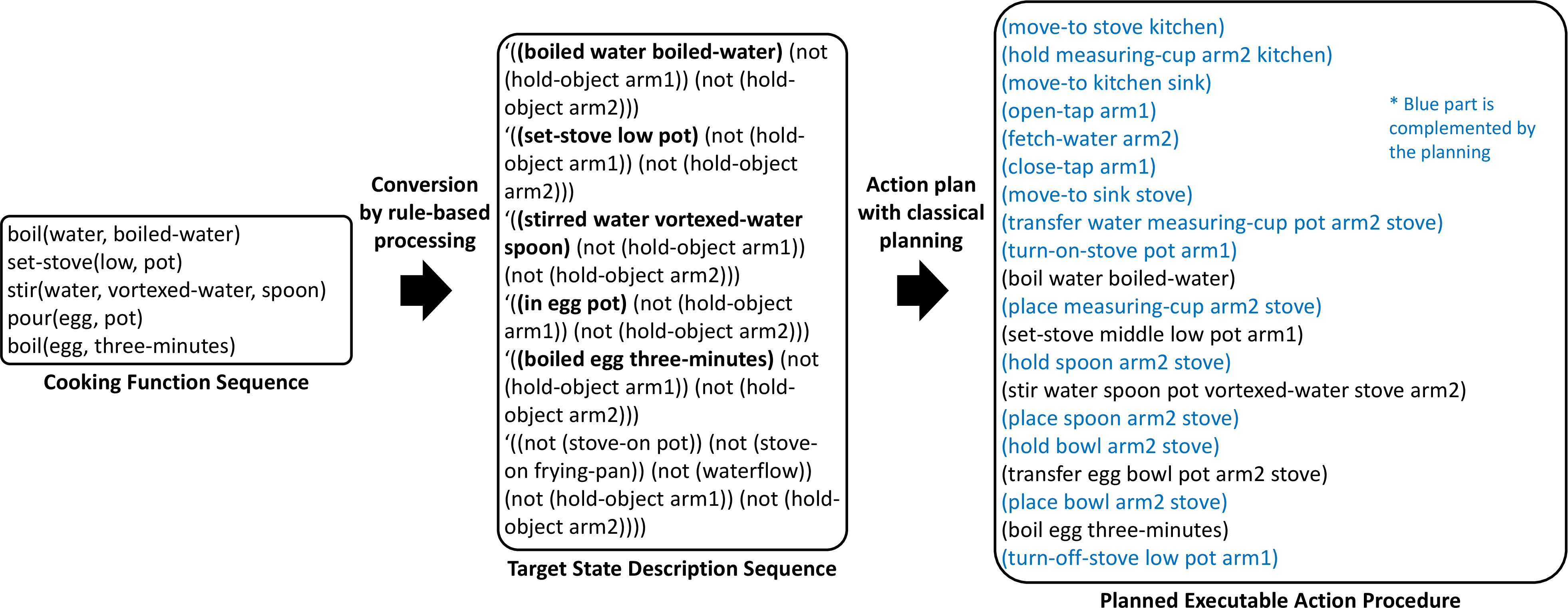}
  \caption{
  Results of executable cooking action procedure planning with PDDL for the poached egg cooking function sequence.
  First, the sequence is converted to target state description sequence by rule-based processing. The description in bold is the part converted from the input cooking function sequence, and the other parts are the descriptions of the default and end conditions added by the process.
  Next, action planning is performed using classical planning based on the initial conditions specified by the user. The black text is the action corresponding to the input cooking function sequence, and the blue text is the action procedure complemented by the plan.
  }
  \label{figure:pddl-res-poached}
\end{figure}

\begin{figure}[h]
  \centering
  \includegraphics[width=0.8\columnwidth]{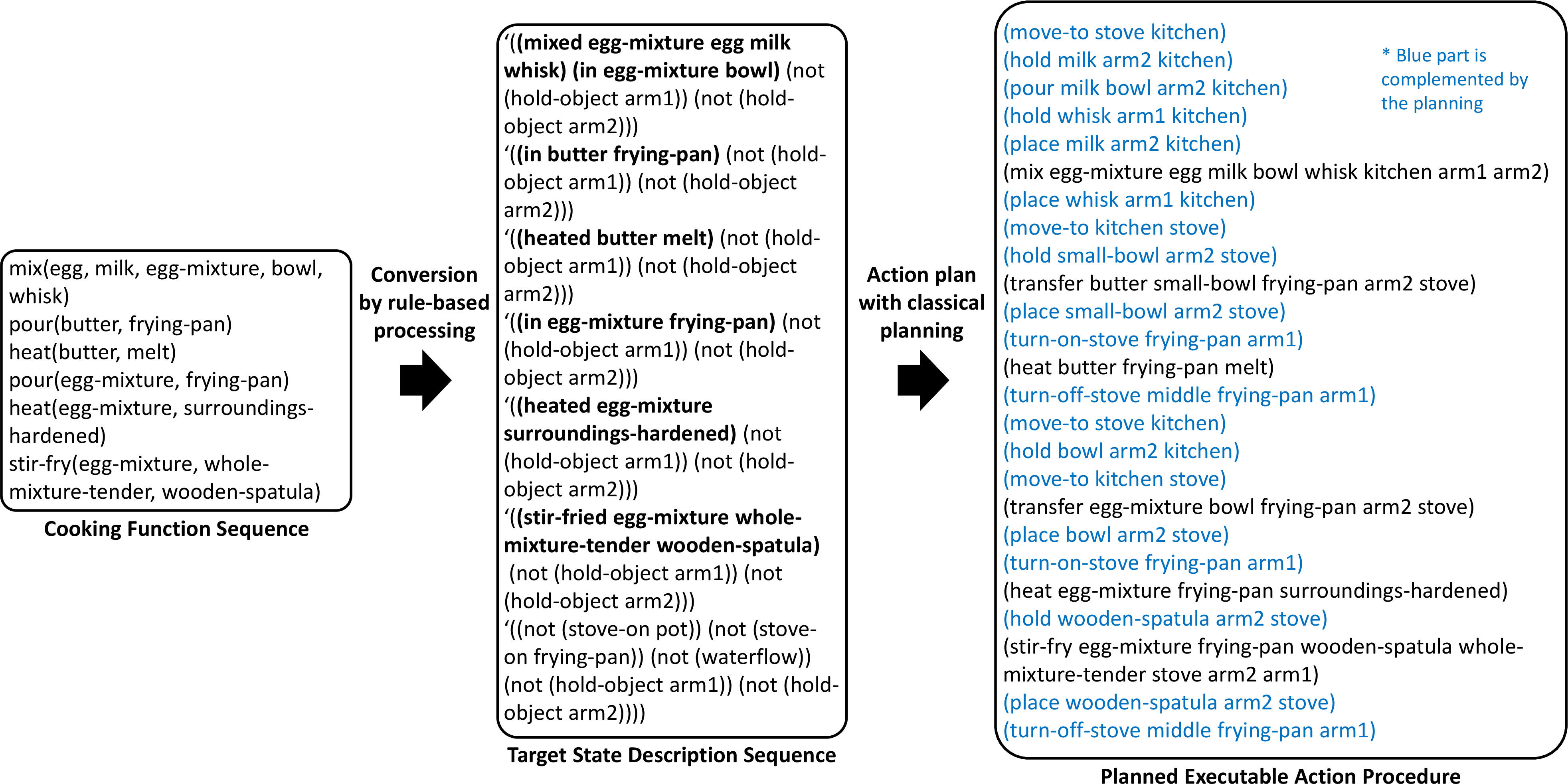}
  \caption{
  Results of executable cooking action procedure planning with PDDL for the scrambled egg cooking function sequence.
  The format of the figure is the same as that of \figref{pddl-res-poached}.
  }
  \label{figure:pddl-res-scrambled}
\end{figure}

\begin{figure}[h]
  \centering
  \includegraphics[width=0.8\columnwidth]{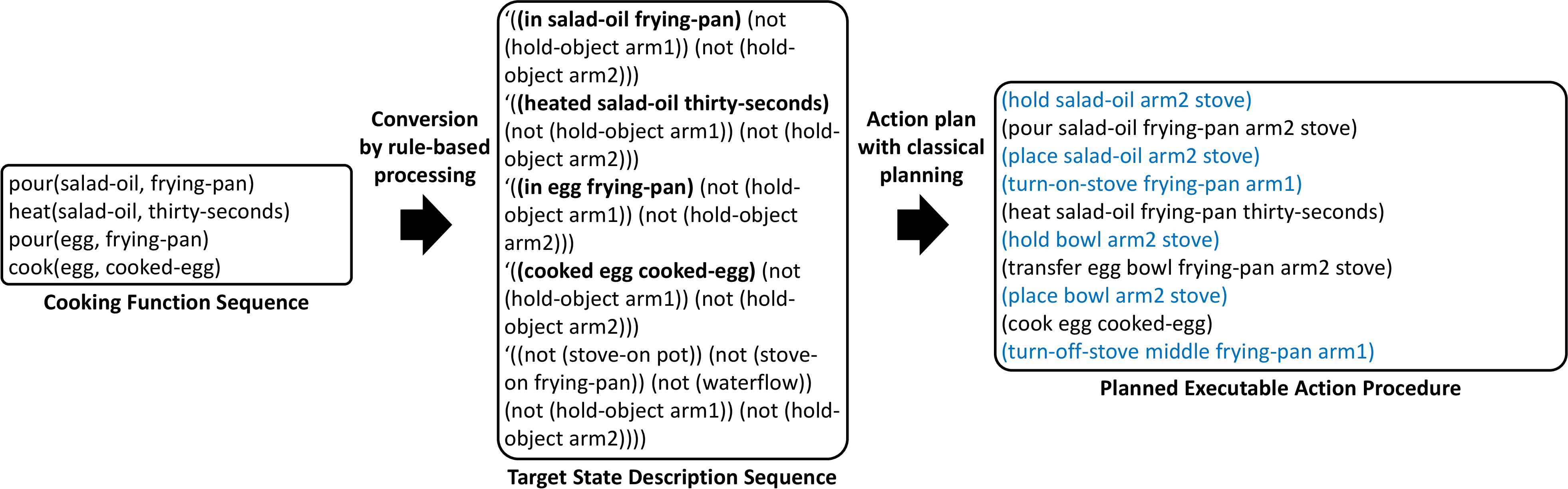}
  \caption{
  Results of executable cooking action procedure planning with PDDL for the sunny-side up cooking function sequence.
  The format of the figure is the same as that of \figref{pddl-res-poached}.
  }
  \label{figure:pddl-res-sunny}
\end{figure}

The result of the action planning for the poached egg is shown in \figref{pddl-res-poached}. Since the initial condition is that there is no water in the pot, the action plan is completed with the following actions: getting a measuring cup, getting water from the tap, pouring water into the pot, and so on.
We also confirmed that the desired action plans for scrambled egg (\figref{pddl-res-scrambled}) and sunny-side up (\figref{pddl-res-sunny}) were also created by complementing the actions such as moving, holding and placing objects, and turning on and off the stove.

\begin{figure}[h]
  \centering
  \includegraphics[width=1.0\columnwidth]{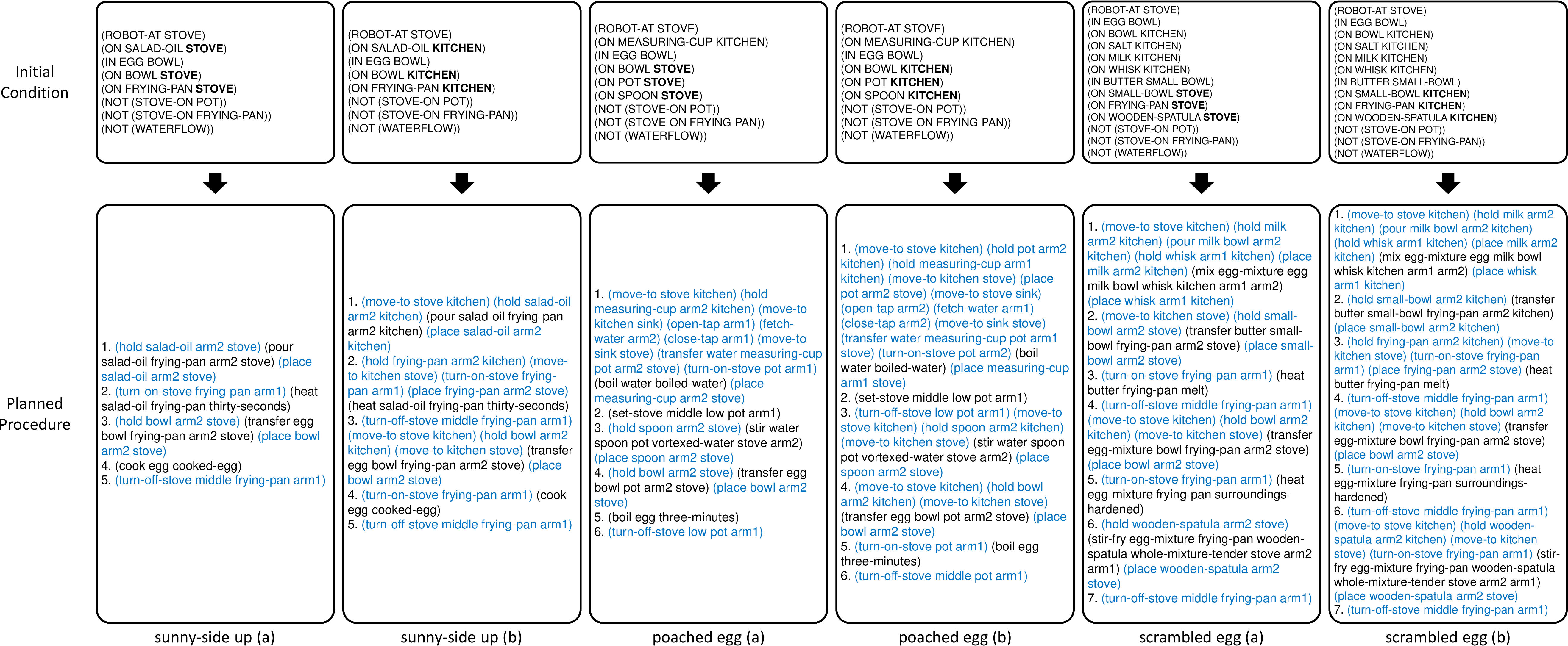}
  \caption{
  Comparison of action plans with PDDL under different initial conditions.
  We compare the action procedure plans in three known recipes under (a) the initial condition in which the person sets so that the number of steps is fewer, as used in experiments such as \figref{pddl-res-poached}, and (b) the initial condition in which all ingredients and utensils are in the kitchen spot. Actions that correspond to one step of the input function sequence are summarized on a single line, with the actions complemented by the planning indicated in blue text and the actions corresponding to the function representation in black text.
  }
  \label{figure:pddl-init-comp}
\end{figure}

An experiment was conducted to plan actions with PDDL under different initial conditions than in the previous experiment (\figref{pddl-init-comp}).
For the three known recipes, we also planned actions for the initial condition in which all the tools and ingredients are in the kitchen spot, and confirmed that the action planning was appropriate, with more movement steps than in the previous examples.

\subsection{Experiments of Food State Recognition Learning from Small Data using Vision-Language Model}\label{sec:rec-experimets} 

To evaluate the performance of the recognizer, we collected data on three food state changes in unknown recipes: ``heat(butter, melt)''(Butter melting), "boil(water, boiled-water)''(Water boiling), and "cook(egg, cooked-egg)''(Egg cooking), and conducted experiments to apply the proposed method (\secref{rec}).
In all experiments, image time series data were collected at 10hz, the rate at which CLIP can reason in real time.

\begin{figure}[h!]
  \centering
  \includegraphics[width=1.0\columnwidth]{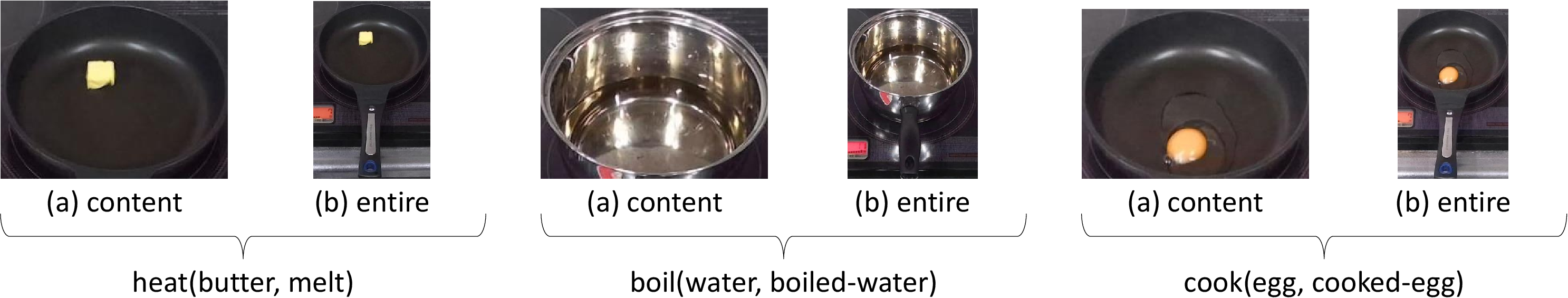}
  \caption{
  Gazing area compared in preliminary experiments.
  The gazing area of only (a) the contents of the pan or pot is compared with the gazing area of (b) the entire pan or pot.
  }
  \label{figure:rec-area}
\end{figure}

\begin{table}[h]
  \centering
  \caption{Results of preliminary experiments to determine the gazing area. The one with the smaller time difference is captioned with bold text.} 
  \scalebox{0.7}{
  \begin{tabular}{|c||c|c|c|c|c|}
    \hline
     & annotation time & estimated time (content) & time difference (content) & estimated time (entire) & time difference (entire) \\
    \hline
    Butter melting & 46.812754926 s & \textbf{38.831157563 s} & \textbf{-7.981597363 s} & 55.672249605 s & 8.859494679 s \\
    Water boiling & 87.728533833 s & \textbf{85.247665755 s} & \textbf{-2.480868078 s} & \textbf{85.247665755 s} & \textbf{-2.480868078 s} \\
    Egg cooking & 115.212714484 s & \textbf{134.72756225 s} & \textbf{19.514847766 s} & 48.9285225 s & -66.284191984 s \\
    \hline
  \end{tabular}}
  \label{table:exep-gaze-area}
\end{table}

First, preliminary experiments on gazing areas were conducted.
As shown in \figref{rec-area}, we compared two cases: (a) a case in which the gazing area is the area of only the contents of the pot or frying pan, and (b) a case in which the gazing area is the entire area of the pot or frying pan.
Two time-series data were obtained for each gazing condition and state change. One data was used as training data to train the recognizer, and state change recognition was performed on the other time-series data.
The performance was evaluated by comparing the inference time of the state change with the annotation time of the person (\tabref{exep-gaze-area}).
``Butter melting'' and ``Egg cooking'' were better predicted by content than by entire, while ``Water boiling'' was predicted at the same time for both.
Since the experiment showed that the gazing area of the content region in (a) gave better results, we decided to adopt the gazing area of the content region from now on.

\begin{figure}[h]
  \centering
  \includegraphics[width=1.0\columnwidth]{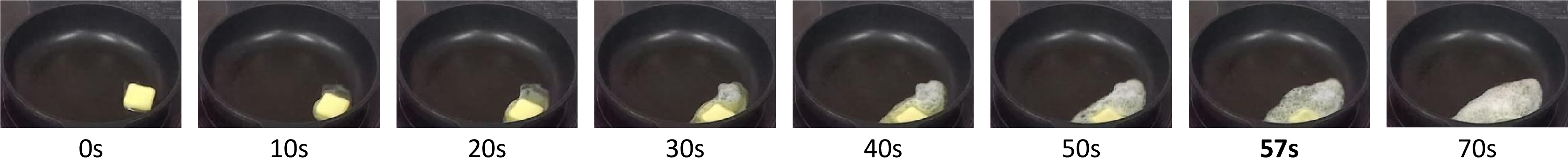}
  \caption{
  Time series data of the food state change of butter melting due to heat.
  The time shown in black bold is the time when the person judged that the state change had occurred.
  }
  \label{figure:state-change-butter}
\end{figure}

\begin{figure}[h]
  \centering
  \includegraphics[width=1.0\columnwidth]{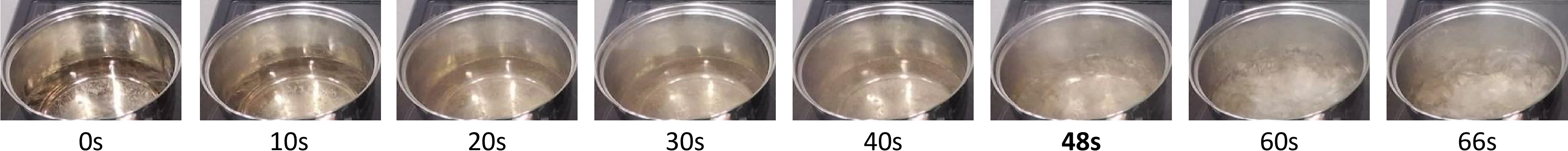}
  \caption{
  Time series data of the food state change of water boiling due to heat.
  The time shown in black bold is the time when the person judged that the state change had occurred.
  }
  \label{figure:state-change-water}
\end{figure}

\begin{figure}[h]
  \centering
  \includegraphics[width=1.0\columnwidth]{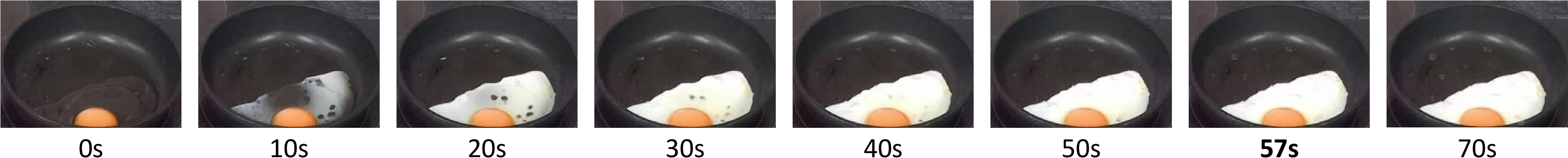}
  \caption{
  Time series data of the food state change of egg cooking due to heat.
  The time shown in black bold is the time when the person judged that the state change had occurred.
  }
  \label{figure:state-change-egg}
\end{figure}

We newly collected data on the three state changes mentioned above, and confirmed that state change recognition was possible.
We obtained four time-series data for each state change, and used the data to train the recognizer and verify its performance. The time-series changes of the used data are shown in \figref{state-change-butter}, \figref{state-change-water}, \figref{state-change-egg}.

\begin{figure}[h!]
  \centering
  \includegraphics[width=1.0\columnwidth]{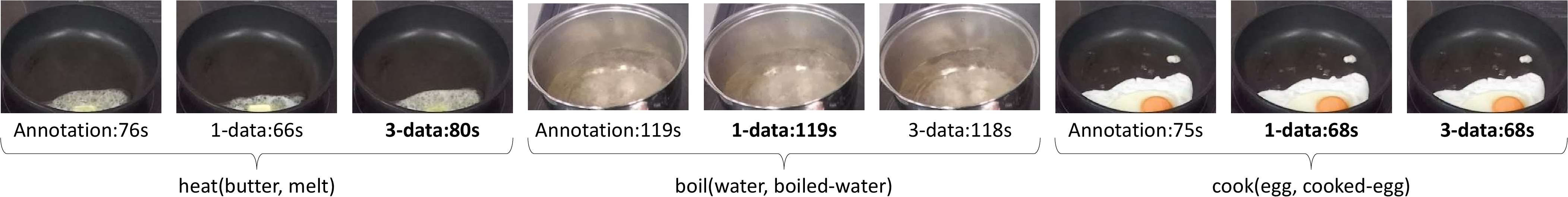}
  \caption{
  Results of food state recognition learning from small data.
  The images at the time of the human annotation and the time at which the respective learning model inferred that the state had changed are shown side by side. The one with the smaller time difference is captioned with bold text.
  }
  \label{figure:rec-result}
\end{figure}

For each of the collected four time series data, three were used for training and one was used for validation.
To test the hypothesis that learning with additional data improves robustness, we compared training with only one of the three training data (1-data) with training with all three (3-data).
The results of comparing the inference time of each state change to the validation data and the time of the person annotations performed beforehand are shown in Table \tabref{exep-rec}.
When evaluated by the smallness of the time difference, 3-data is better for butter melting, 1-data is better for water boiling, and the same number of seconds for egg cooking. Therefore, it is not necessarily true that more training data is better.
However, a qualitative comparison including images at the estimated time of \figref{rec-result} yields a different result.
In the 1-data of butter melting, the time difference is large (more than 10 seconds) and the butter is not yet melted in the image, while in the 3-data of water boiling, it is difficult to say from the image condition that boiling has not yet occurred, and the time difference is relatively small (about 1 second).
Therefore, the 3-data error is smaller in magnitude than the 1-data error, and it is considered that a better recognizer can be learned by increasing the number of training data.
Therefore, in the real robot experiment (\secref{cook-experimets}) described below, we use the 3-data recognizer.
\begin{table}[h]
  \centering
  \caption{Results of food state recognition learning from small data. The one with the smaller time difference is captioned with bold text.} 
  \scalebox{0.7}{
  \begin{tabular}{|c||c|c|c|c|c|}
    \hline
     & annotation time & estimated time (1-data) & time difference (1-data) & estimated time (3-data) & time difference (3-data) \\
    \hline
    Butter melting & 76.063834997 s & 65.613179821 s & -10.450655176 s & \textbf{79.900955305 s} & \textbf{3.837120308 s} \\
    Water boiling & 118.754276608 s & \textbf{118.976917507 s} & \textbf{0.222640899 s} & 117.650410837 s & -1.103865771 s \\
    Egg cooking & 75.024540634 s & \textbf{68.327492236 s} & \textbf{-6.697048398 s} & \textbf{68.327492236 s} & \textbf{-6.697048398 s} \\
    \hline
  \end{tabular}}
  \label{table:exep-rec}
\end{table}

\subsection{Real-World Experiments with Robot Cooking Execution from Recipes}\label{sec:cook-experimets} 

\begin{figure}[h]
  \centering
  \includegraphics[width=1.0\columnwidth]{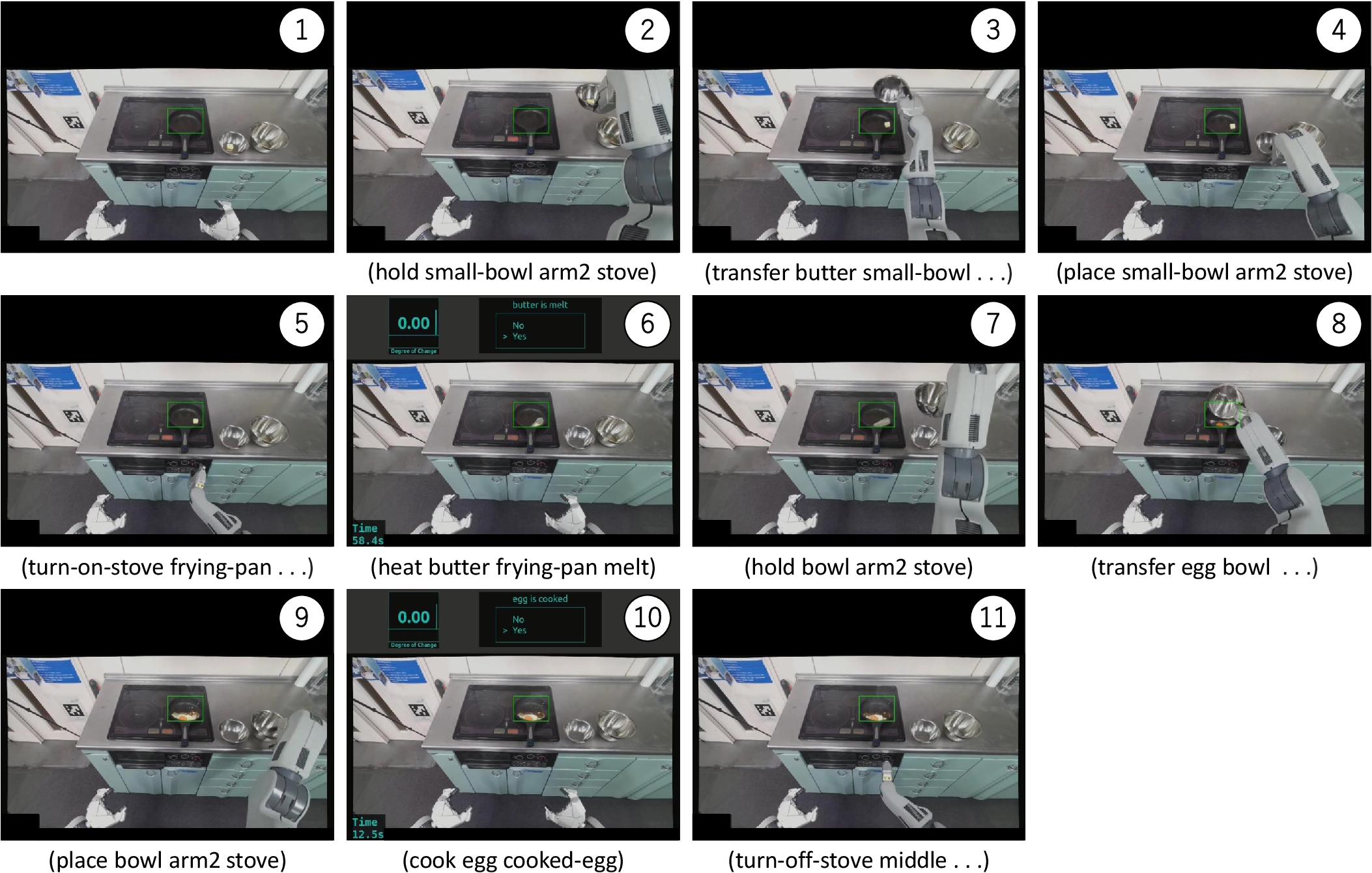}
  \caption{
  Cooking execution experiment from butter arranged sunny-side up recipe.
  It was confirmed that the proposed system can actually perform a series of sunny-side up cooking based on a natural language recipe description, with action planning and ingredient state recognition.
  }
  \label{figure:experiment-butter}
\end{figure}

\begin{figure}[h]
  \centering
  \includegraphics[width=1.0\columnwidth]{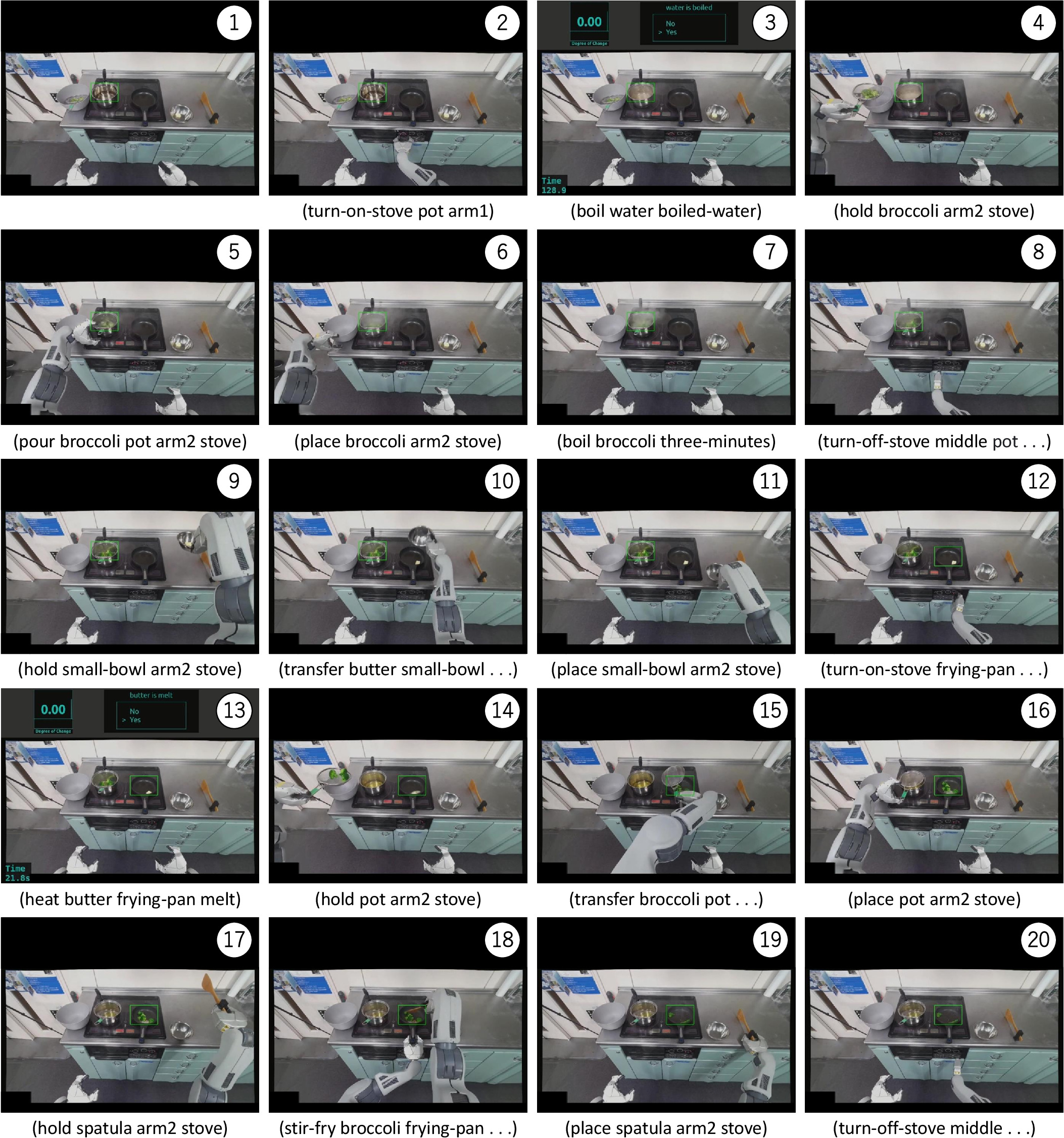}
  \caption{
  Cooking execution experiment from boiled and sauteed broccoli recipe.
  It was confirmed that the proposed system can actually perform a series of boiled and sauteed broccoli cooking based on a natural language recipe description, with action planning and ingredient state recognition.
  }
  \label{figure:experiment-broccoli}
\end{figure}

Experiments (\figref{experiment-butter} and \figref{experiment-broccoli}) were conducted in which the robot cooked an arranged unknown recipe in the real world, and the effectiveness of the proposed system was confirmed.
The proposed system plans cooking actions that can be executed in the real world based on natural language recipe descriptions (\secref{plan}), and executes them in sequence while recognizing changes in the state of food ingredients during heating using the Vision-Language model(\secref{rec}).
In this experiment, the robot performed the motions that were created by a human by using direct teach, etc.
Please refer to the attached movie for the actual cooking process as seen from the robot's camera.

\begin{figure}[h]
  \centering
  \includegraphics[width=0.8\columnwidth]{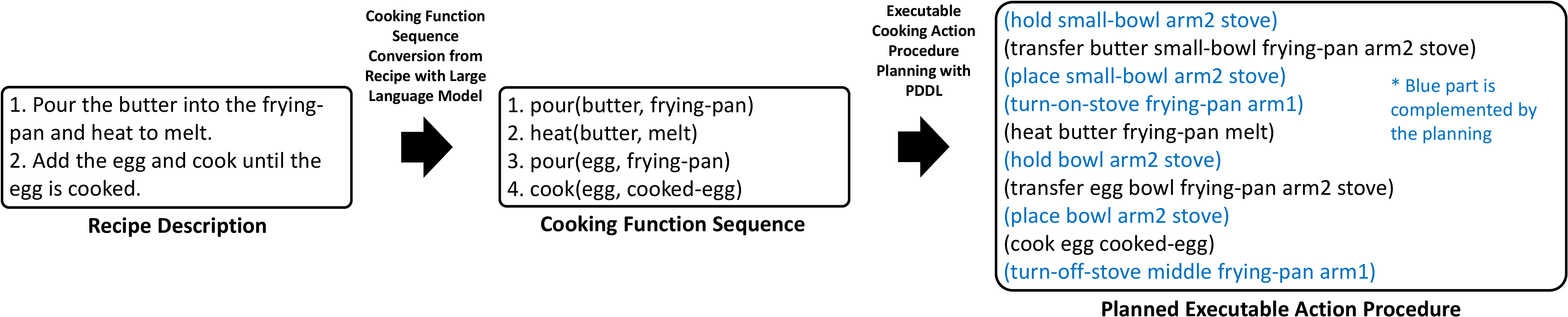}
  \caption{
  The result of the overall action plan for an unknown recipe of sunny-side up arranged to use butter.
  The action procedure for the prepared unknown recipe was planned by conversion by rule-based processing and action planning with classical planning.
  The blue part of the planned executable action procedure is the action complemented by the PDDL planning.
  }
  \label{figure:unknown-butter}
\end{figure}

A new recipe for sunny-side up, arranged to use butter instead of oil, was cooked by the robot in the real world using the proposed system.
First, as shown in \figref{unknown-butter}, the robot planned actions that could be executed in the real world from the given natural language recipe using the method proposed in \secref{plan}, converting it into a cooking function sequence using LLM, and then planning PDDL from the sequence.
Since the starting condition was a situation in which butter and eggs were placed in bowls, actions such as holding the bowls containing the butter and eggs and turning on the stove were complemented.

The robot executed the planned cooking actions while recognizing the state of the food ingredients using the method proposed in \secref{rec}.
The robot experimenting with cooking autonomously is shown in \figref{experiment-butter}.
The robot's recognition of whether the eggs were cooked or not ended immediately because the frying pan was quite hot when the eggs were poured and it took time to put the bowl down after the eggs were poured.
However, when the sunny-side up was transferred from the frying pan to a plate immediately after the robot finished cooking, it was confirmed that the sunny-side up was cooked properly without collapsing.

\begin{figure}[h]
  \centering
  \includegraphics[width=0.8\columnwidth]{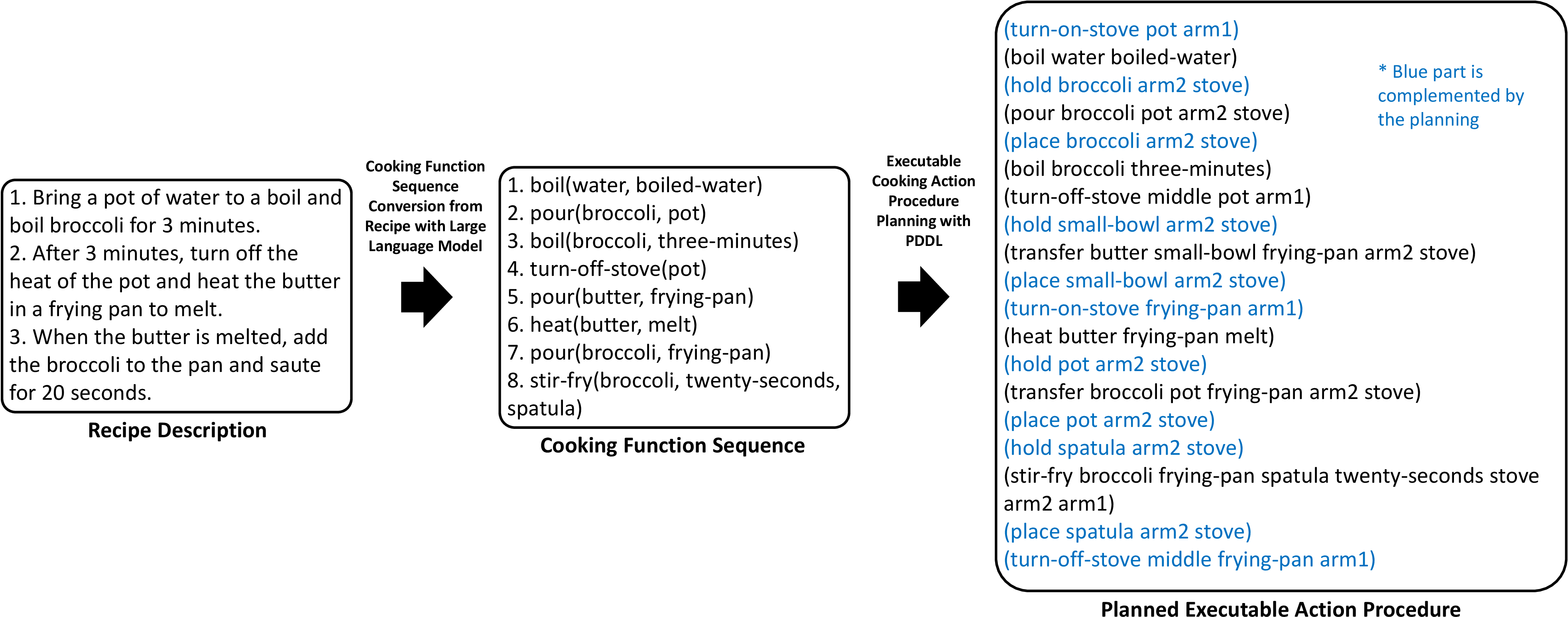}
  \caption{
  The result of the overall action plan for an unknown recipe of boiled and sauteed broccoli.
  The format of the figure is the same as that of \figref{unknown-butter}.
  }
  \label{figure:unknown-broccoli}
\end{figure}

The cooking was executed in the same way for the unknown recipe of boiled and sauteed broccoli.
Real-world action planning from the recipe was performed as shown in \figref{unknown-broccoli}.
The robot autonomously executes the planned action steps while recognizing changes in the state of the food ingredients as shown in \figref{experiment-broccoli}.
Although there are some points to be improved, such as the fact that the robot uses a net ladle for the experimental circumstance, the real-world behavior is a little irregular in this part, and the time until the robot judges that the butter is melted is short, and it seems that the butter was not yet completely melted when the robot judged.
However, we were able to actually cook broccoli.

\begin{figure}[h]
  \centering
  \begin{minipage}{0.24\columnwidth} 
    \centering
    \includegraphics[width=1.0\columnwidth]{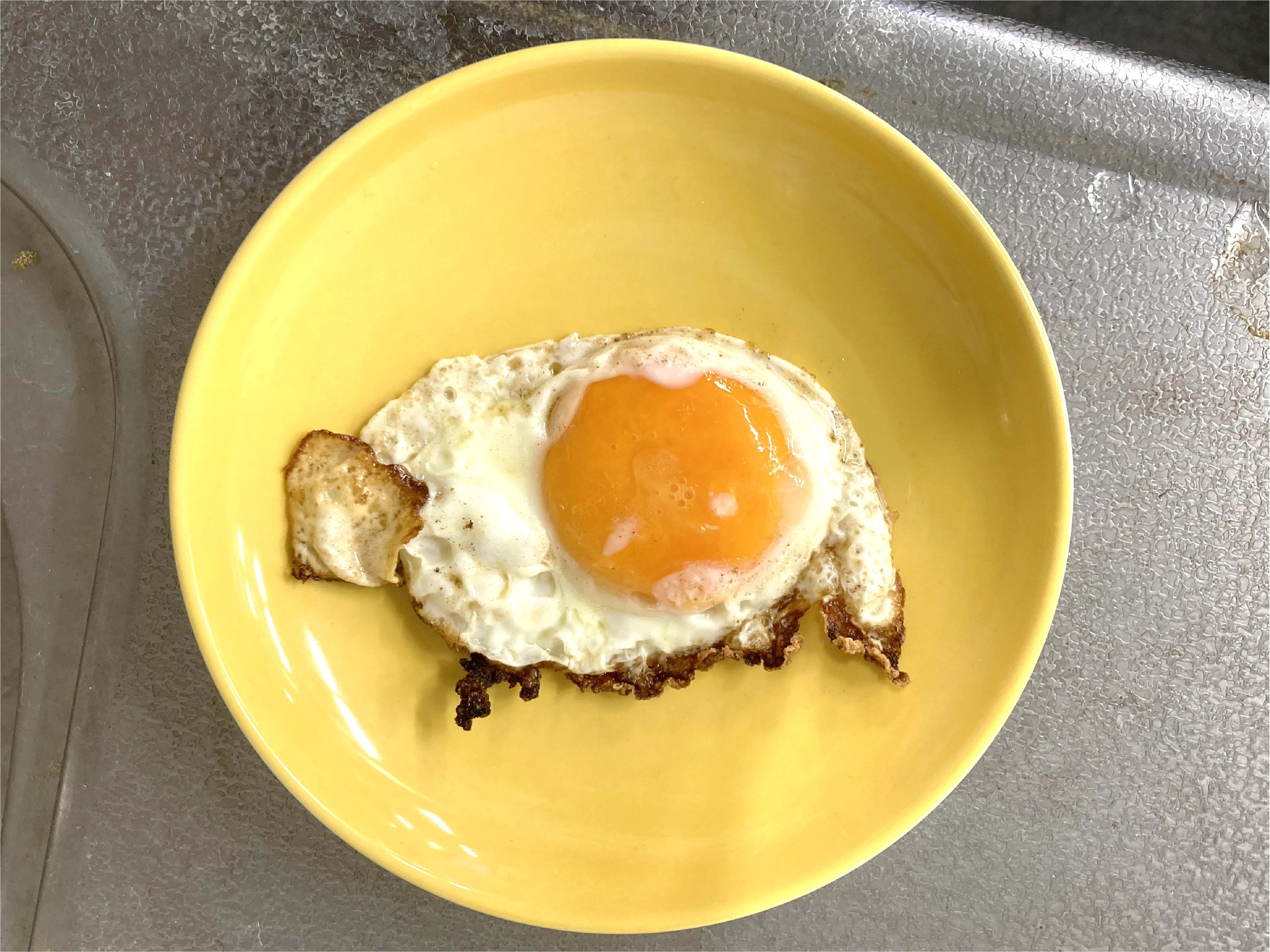}
    \subcaption{Butter arranged sunny-side up}
  \end{minipage}
  \begin{minipage}{0.24\columnwidth}
    \centering
    \includegraphics[width=1.0\columnwidth]{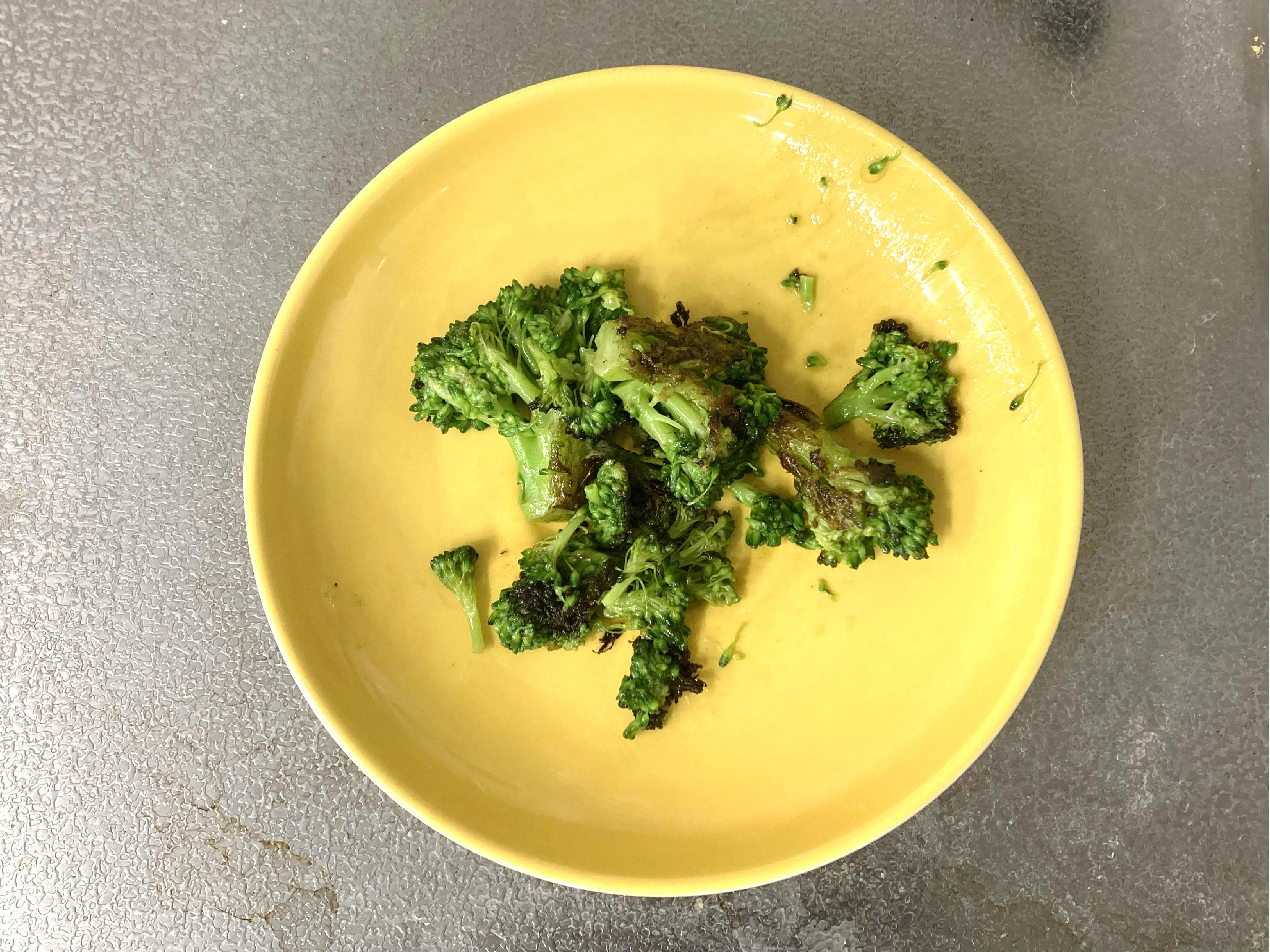}
    \subcaption{Boiled and sauteed broccoli}
  \end{minipage}
  \caption{Robot cooked dishes. We confirmed that the finished dishes were tasty and human-eatable by eating them.}
  \label{figure:experiment-result-dish}
\end{figure}

The robot's cooked dishes served on plates are shown in \figref{experiment-result-dish}.
The robot executed the cooking using real ingredients in the real world, and confirmed that the finished dishes were tasty and human-eatable by eating them.

\section{Discussion}

This section mainly describes the performance and limitations of this study.
First, the action plan part is described. The range of recipes that can be covered by this setup is limited due to the small number of cooking functions prepared, and it is considered possible to handle cocktails, smoothies and other juices, salads, soups, and other dishes on the assumption that the ingredients have been prepared.
In order to handle more recipes, it is necessary to add more cooking functions, and although our method can handle complex recipes with many steps, further efforts are needed to keep the accuracy of the action plan high even when the recipes are complex.
In interpreting complex recipes, it may be difficult to maintain continuity of the object indicated by the function arguments between steps of the function sequence.
It would be effective to express the arguments as variables with indices and check whether the entire functional expression is convertible into a valid graph by referring to the previous study on the function expression of cooking~\cite{otasse2008sour, jermsurawong2015predicting}, and to re-plan by feeding back error messages and other results to the LLM by referring to the previous study on the use of LLM~\cite{shirai2023vision}.
In addition, in cooking, it is possible to execute multiple tasks simultaneously using multiple stoves, such as simultaneous cooking of multiple dishes or simultaneous cooking of pasta sauce and noodles, and PDDL itself can also execute multiple tasks simultaneously.
Therefore, the action planning method combining LLM and PDDL in this study is expected to be effective in the future development of cooking robot systems, including more complex recipes and simultaneous execution of multiple tasks.

Next, we will discuss the part of the recognition of state change of food ingredients.
In cooking, various state changes occur, and it is necessary to recognize state changes in situations where there are multiple ingredients in a pot or pan. Our state recognition method is able to deal with such situations by learning from a small amount of the relevant data. For situations where there is no data for the same situation, there is a possibility that state recognition for unknown state changes and unknown situations can be improved by referring to the research on generalization of state change recognition to different objects~\cite{xue2024learning} for further improvement, which should be verified in the future.
On the other hand, the method in this study is also considered to be very effective for stable state change recognition in the presence of a small number of data, and a system that uses two different methods for unknown and known states may be effective.
It will be also important to recognize whether the food has become soft or not by tactile sensation, to sense the state of burnt food and temperature by sense of smell and temperature, and to evaluate the final dish by these senses together with taste.

Finally, we discuss the motion execution part.
In this research, cooking was performed by executing motion trajectories predefined by a human.
Finally, we discuss the motion execution part.
In this research, cooking was performed by executing motion trajectories predefined by humans. However, in order for robots to be able to perform more complex movements and manipulation of foods with individual differences, it will be necessary to implement adaptive motion execution using machine learning.
Efficient use of imitation learning~\cite{chi2023diffusion}, which learns movement skills from human piloted data, in combination with image recognition or visual information extraction using the Foundation Models, etc., is considered effective.

\section{Conclusion}

In this study, we proposed a robotic system in which robots perform real-world cooking based on food recipes written in natural language.
For action planning, we used the Large Language Model (LLM) and classical planning of PDDL descriptions to plan robot cooking actions that can be executed in the real world.
The LLM's few-shot prompting transformed a natural language recipe into a sequence of cooking functions that can be interpreted by robots.
By using PDDL planning based on the converted sequences, we plan action procedures that can be executed in the real world from environment- and agent-independent recipe descriptions.
For recognition, we learned to recognize the state of food ingredients from a small amount of data using the Vision-Language Model (VLM) to realize real-time recognition of food state changes during cooking.
Experiments confirmed that the accuracy of the recognizer increases when the region of the contents of the pot or frying pan is used as the gazing area, rather than the entire region of the pot or frying pan, and that the accuracy of the recognizer increases when more time-series data is used for training, rather than training from a single time-series data set.
We constructed a system that integrates these two methods, and conducted experiments in which a robot cooked food autonomously from unknown recipes in the real world, and confirmed the effectiveness of the proposed system.

In the future, we will extend the overall system to be able to execute more dishes and recipe descriptions in various environments by performing motion planning based on recipe descriptions and integrating the whole cooking execution robot system.
The cooking function representation part may be extended to be able to handle more food categories, but the action planning and ingredient state recognition proposed in this research will be effective in that case as well.







{
\bibliographystyle{unsrt}
\bibliography{main}
}

\end{document}